\documentclass[10pt,twocolumn,letterpaper]{article}

\usepackage{iccv}              
\usepackage{multirow} 

\definecolor{iccvblue}{rgb}{0.21,0.49,0.74}
\usepackage[pagebackref,breaklinks,colorlinks,allcolors=iccvblue]{hyperref}

\def\paperID{12180} 
\def\confName{ICCV}
\def\confYear{2025}

\title{SPADE: Spatial-Aware Denoising Network for Open-vocabulary Panoptic Scene Graph Generation with  Long- and Local-range  Context Reasoning}

\author{Xin Hu$^1$, Ke Qin$^{1,2}$, Guiduo Duan$^{1,2}$, Ming Li$^4$, Yuan-Fang Li$^3$, Tao He$^{1}$ \thanks{Corresponding author.}\\
$^1$The Laboratory of Intelligent Collaborative Computing of UESTC \\
$^2$Ubiquitous Intelligence and Trusted Services Key Laboratory of Sichuan Province \\
$^3$ Monash University, 
$^4$ Guangdong Laboratory of Artificial Intelligence and Digital Economy (SZ) \\
{\tt\small \{xh1m22,qinke,guiduo.duan\}@uestc.edu.cn}\\
{\tt\small ming.li@u.nus.edu, yuanfang.li@monash.edu, tao.he01@hotmail.com}
}

\begin{document}
\maketitle
\begin{abstract}
Panoptic Scene Graph Generation (PSG) integrates instance segmentation with relation understanding to capture pixel-level structural relationships in complex scenes. Although recent approaches leveraging pre-trained vision-language models (VLMs) have significantly improved performance in the open-vocabulary setting, they commonly ignore the inherent limitations of VLMs in spatial relation reasoning, such as difficulty in distinguishing object relative positions,  which results in suboptimal relation prediction.
Motivated by the denoising diffusion model's inversion process in preserving the spatial structure of input images, we propose \textbf{SPADE} (\underline{SP}atial-\underline{A}ware \underline{D}enoising-n\underline{E}twork) framework---a novel approach for open-vocabulary PSG. SPADE consists of two key steps: (1) inversion-guided calibration for the UNet adaptation, and (2) spatial-aware context reasoning. In the first step, we calibrate a general pre-trained teacher diffusion model into a PSG-specific denoising network with cross-attention maps derived during inversion through a lightweight LoRA-based fine-tuning strategy. In the second step, we develop a spatial-aware relation graph transformer that captures both local and long-range contextual information, facilitating the generation of high-quality relation queries.
Extensive experiments on benchmark PSG and Visual Genome datasets demonstrate that SPADE outperforms state-of-the-art methods in both closed- and open-set scenarios, particularly  for spatial relationship prediction. The project page is at \href{https://8078qwe.github.io/SPADE/}{here}.

\end{abstract}

\section{Introduction}
\label{sec:intro}

Panoptic Scene Graph Generation (PSG) represents a significant advancement in scene graph generation (SGG) by integrating pixel-level structural analysis with relation reasoning to achieve comprehensive scene understanding. By unifying instance segmentation \cite{ren2015faster,Liu_2018_CVPR} and visual relationship detection \cite{chao2015hico,gao2018ican} within a single framework, PSG attempts to simultaneously segment all entities in an image and classify their relational interdependencies into subject–predicate–object triples.

In prior research, the majority of  SGG  \cite{hayder2024dsgg, cong2023reltr, he2021exploiting, li2022sgtr, he2021learning} and  PSG  methods \cite{yang2022panoptic, zhou2023hilo, wang2024pair} have been developed from a closed-set perspective. 
Recent advancements, however, have introduced an open-vocabulary setting for SGG \cite{zareian2021open, he2022towards},  shifting research focus toward a more practical yet challenging problem. Within this paradigm, various methods \cite{chen2023expanding, zhou2024openpsg, li2024pixels, yu2023visually, zhang2023learning} have been proposed to endow their models with open-world recognition capabilities. A key technique in these works is to leverage the powerful pre-trained Vision Language Models (VLMs) \cite{clip, li2022grounded, li2023blip}, enabling the recognition of unseen objects and relationships.

\begin{figure}[t]
\centering
\includegraphics[width=0.5\textwidth]{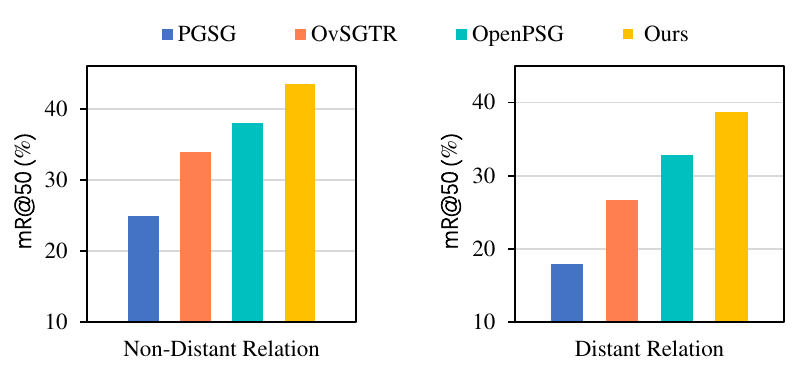}
\caption{
Mean Recall@50  of our proposed
method and other VLM-based  models \cite{li2024pixels, chen2023expanding, zhou2024openpsg} on the spatial  predicate classification task on the PSG dataset \cite{yang2022panoptic}. Notably, we present the result in the two scenarios: distant and non-distant relation pairs. 
}
\label{fig:intro}
\end{figure}

 
However, a suite of recent studies \cite{spatialvlm,cheng2025spatialrgpt,li2024topviewrs,xu2023open,shiri-etal-2024-empirical} have revealed that most state-of-the-art VLMs, such as CLIP \cite{clip},  BLIP \cite{li2023blip}, and  GLIP \cite{glip},  struggle with spatial reasoning, particularly in understanding spatial relationships between objects. This limitation primarily arises because such VLMs are trained on internet-scale datasets composed of image-caption pairs, which often lack detailed spatial descriptions. Consequently, this raises concerns about whether VLM-based  SGG  or PSG  models can still perform effectively in spatial relation prediction.  To address this issue, we conduct a comprehensive investigation of several state-of-the-art VLM-based open-set PSG and SGG models, with a specific focus on their performance in spatial relation prediction. Detailed experimental protocols are provided in \S \ref{spatial}. The results, as illustrated in Figure \ref{fig:intro}, indicate that these VLM-based models underperform on the spatial relation, particularly when two objects are positioned at a considerable distance from each other within an image. We conjecture that increased inter-object distance complicates the VLMs to  determine spatial relationships.

Motivated by these findings, we pose a question: \emph{Can we integrate spatial knowledge into a VLM without compromising its inherent open-world recognition capabilities for PSG?} To address this question, we turn to diffusion models for two main reasons. First, diffusion models \cite{rombach2022high, balaji2022ediff, saharia2022photorealistic} exhibit remarkable compositional abilities that preserve both spatial and semantic consistency, enabling the injection of spatial knowledge without degrading other semantic information as much as possible. Second, recent work \cite{gddim,wang2025belm} has demonstrated that the inversion process \cite{ddim}, widely utilized in image editing \cite{wang2023stylediffusion,kawar2023imagic}, effectively preserves the spatial structure of original images. Thus, we believe that the cross-attention maps derived from the inversion process can serve as explicit spatial cues.  However, directly leveraging the pretrained diffusion model does not yield the desired performance, as its preserved knowledge is not fully adapted to the  PSG task.

Based on the insights, we present \textbf{SPADE} (\underline{SP}atial-\underline{A}ware \underline{D}enoising-n\underline{E}twork), a novel two-stage approach for PSG. The proposed method addresses the challenges of spatial relation reasoning and open-vocabulary generalization through two key steps: (1) inversion-guided calibration for the UNet adaptation and (2) spatial-aware context reasoning via graph-based relation inference.
In the first step,  our core idea is to adapt a general pre-trained teacher diffusion model into a PSG-specific denoising network.  Building on the observation that the Denoising Diffusion Implicit Models (DDIM) inversion  \cite{wang2025belm,ddim} inherently retains spatial information from input images, we leverage cross-attention maps derived during inversion as spatial priors.  To preserve the generalization capabilities of the pre-trained UNet, we leverage a lightweight fine-tuning strategy LoRA \cite{hu2021lora} with only updating cross-attention layers.

In the second step, SPADE subsequently employs a Relation Graph Transformer (RGT) to capture long-range contextual dependencies and local spatial cues. The RGT operates on a spatial-semantic graph where nodes represent object mask proposals with geometric attributes and semantic embeddings. Through iterative graph propagation, the transformer layer aggregates multi-scale contextual features, resolving ambiguities in overlapping relations and sparse spatial configurations. This dual-context mechanism—combining long-range and local correlations—enables to generate high-quality relation queries.

To summarize, our contributions are as follows:
\begin{itemize}
    \item We identify a shortcoming in the spatial relation prediction  of current VLM-based   PSG models. To address this limitation, we propose a two-step spatial-aware denoising network (\textbf{SPADE}).
    
    \item We introduce an inversion-guided calibration strategy that utilizes cross-attention maps from the reversion process to fine-tune the  UNet in a LoRA fashion. 
    \item We design a spatial-aware relation graph transformer that captures local and long-range pairwise contextual information, enabling robust spatial relation reasoning.
    
    \item Extensive experiments on benchmark datasets such as VG and PSG demonstrate that \textbf{SPADE} significantly outperforms state-of-the-art methods in both closed-set and open-set scenarios, particularly in improving relation prediction in open-vocabulary settings.
\end{itemize}


\section{Related Work}






\noindent\textbf{Open-world Panoptic Scene Graph Generation.} Panoptic Scene Graph Generation (PSG)  or Scene Graph Generation (SGG)   aims to represent an image through structured relationships among detected objects \cite{krishna2017visual,yang2022panoptic}. Traditional methods generally focus on a closed set of categories, which limits their capacity to generalize to novel objects and relationships \cite{chiou2021recovering, he2023toward, zellers2018neural,tang2019learning,he2021semantic}. Recently, open-world SGG has emerged to address this challenge \cite{zareian2021open,chiou2021recovering, xu2017scene, zellers2018neural,tang2019learning,kan2021zero,yu2022zero}. Earlier zero-shot approaches often leverage commonsense knowledge or structured knowledge graphs to transfer understanding from known to unknown relations \cite{kan2021zero,yu2022zero}. Later advancements
integrate multi-modal models \cite{yu2023visually,zhang2023learning}. However, 
they commonly ignore the limitation of VLMs in spatial relation reasoning, which hinders them to predict accurate spatial relationships. In this work, we turn to  diffusion models \cite{rombach2022high}  and imbue the spatial knowledge into the denoising network, enhancing its spatial   reasoning ability.   



\noindent \textbf{DDIM Inversion Process.} Denoising Diffusion Implicit Models (DDIM) have recently demonstrated remarkable success in generating high-quality images conditioned on text \cite{rombach2022high,balaji2022ediff,wang2025belm,saharia2022photorealistic}. These models operate via a forward process that progressively adds Gaussian noise to an image, and a reverse process that denoises the image. Formally, the forward (noising) process is defined as:
\begin{equation}
    x_t = \sqrt{\bar{\alpha}_t} \, x + \sqrt{1-\bar{\alpha}_t} \, \epsilon, \quad \epsilon \sim \mathcal{N}(0,I),
\end{equation}
where \(\bar{\alpha}_t = \prod_{k=1}^{t}\alpha_k\). The inversion process maps an input image \(x\) into a deterministic noise. During deterministic sampling \cite{ddim}, the image is reconstructed from the noise:
\begin{equation}
\begin{split}
    x_{t-1} = \sqrt{\bar{\alpha}_{t-1}} \left(\frac{x_t - \sqrt{1-\bar{\alpha}_t} \, \epsilon_\theta(x_t, t)}{\sqrt{\bar{\alpha}_t}}\right) \\
    + \sqrt{1-\bar{\alpha}_{t-1}} \, \epsilon_\theta(x_t, t),
\end{split}
\end{equation}
where \(\epsilon_\theta(x_t, t)\) denotes the predicted noise.  In this work, the cross-attention maps derived from deterministic sampling are exploited to extract rich spatial and semantic priors. These priors are instrumental in guiding the calibration of the denoising network, thereby enhancing its suitability for open-world panoptic scene graph generation.

\noindent \textbf{Graph Transformer.}
Graph transformers have emerged as a powerful paradigm for modeling node relationships by effectively capturing both global context and local dependencies within structured data \cite{shehzad2024graph,shaw2018self,yun2019graph}. This makes it well-suited for reasoning computer vision tasks, where understanding complex object relationships is crucial \cite{muller2023attending,xia2021chief}.   This   approach not only enhances the modeling of object interactions but also contributes to improved overall scene understanding \cite{sortino2023transformer,cong2023reltr}. In this work, we capitalize on these strengths by integrating graph transformers into our framework to bolster relationship reasoning. Through this integration, our model is better equipped to capture intricate inter- and intra-object dependencies.

\begin{figure*}[t]
  \includegraphics[width=1\textwidth]{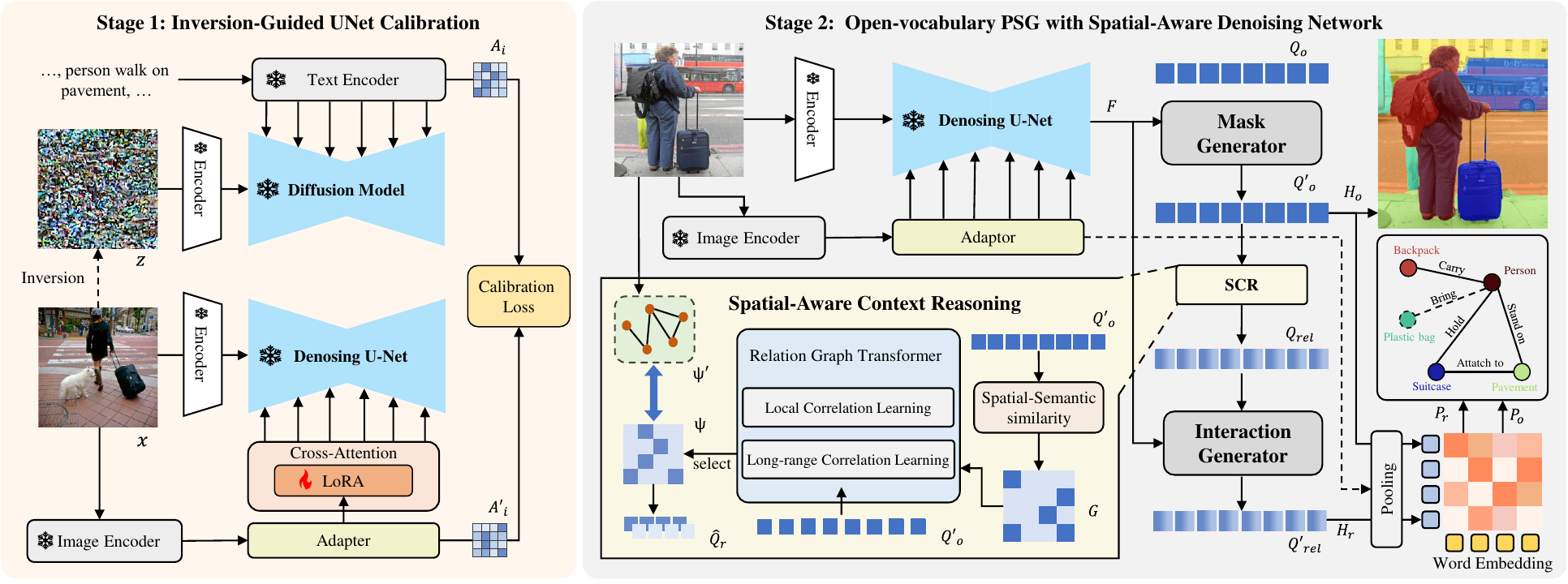}
  \centering
  \caption{ The overview of \textbf{SPADE}, which comprises two key steps: (1) inversion-guided calibration for the UNet to adapt a general pre-trained teacher diffusion model into a PSG-specific denoising network and (2) {spatial-aware context reasoning} (SCR) over relativeness and non-relativeness context   through a relation graph Transformer to generate high-quality relation queries.}
  
  \label{fig:frame}
\end{figure*}

\section{Methodology}


The overall architecture of \textbf{SPADE} is illustrated in Figure \ref{fig:frame}. Briefly, SPADE is composed of two primary stages:  inversion-guided calibration and spatial-aware relation context inference. 
The first stage focuses on imbuing the pre-trained denoising network with spatial and relation knowledge (\S \ref{sec:4.1}), while the second stage, built on the calibrated UNet, performs spatial-aware context reasoning by a novel relation graph transformer (\S \ref{sec:4.2}).   Lastly, we present how we make the open-vocabulary relation prediction (\S \ref{sec:4.3}).  


\subsection{Inversion-Guided Calibration for the UNet}\label{sec:4.1}
Diffusion models have emerged as backbones for many downstream vision tasks \cite{xu2023open,kondapaneni2024text,yang2024open,li2023open,zhao2023unleashing}. Formally, given an input image-text pair \((x, w)\), these models first construct a noisy image \(x_t\) and encode the text \(w\) using a pre-trained text encoder \(\epsilon_t\). Intermediate features \(f_t\) are then extracted from the UNet as follows:
\begin{equation}
    f_t = \mathrm{UNet}(x_t, \epsilon_t(w)).
\end{equation}
In the context of PSG, however, an input text description \(w\) is not available at inference. Following \cite{yang2024open,xu2023open}, one option is to replace \(\epsilon_t(w)\) with a global image feature obtained from the CLIP \cite{clip} image encoder.

Nevertheless, we observe that this approach does not yield satisfactory performance for relation prediction. We hypothesize two main reasons for that: (1) although the alternative CLIP  features effectively capture object- and pixel-level information, they are less proficient at discerning spatial and scene-level relation \cite{xu2023open,subramanian2022reclip}; and (2) the  diffusion model is trained using long, context-rich descriptions, which may not align well with relation triple phrases \cite{zhao2023unleashing,kondapaneni2024text,li2023open}, resulting in incompatibility with the PSG task.

To overcome these limitations, we introduce an inversion-guided calibration approach for the denoising network, leveraging the inversion process \cite{ddim,gddim,wang2025belm}, which has been shown to effectively preserve spatial information from input images. Our calibration process aims to adapt a pre-trained diffusion model into a PSG-specific UNet by explicitly injecting spatial and relation knowledge. 

\noindent \textbf{Inversion for Denoised Feature Extraction.}
Inversion techniques \cite{ddim,gddim,wang2025belm} have been widely adopted in various applications, such as image editing \cite{wang2023stylediffusion,kawar2023imagic}, owing to their ability to preserve the spatial information inherent in the original image. Leveraging this property, we first forward  realistic images \( x \) (e.g., from the VG \cite{krishna2017visual}) and obtain corresponding deterministic noise representations \( z \) through the inversion process. Next, we design a relation prompt of the form 
\(
p = \texttt{[subject] is [predicate] [object]...},
\)  
where each placeholder (e.g., \texttt{[subject]}) corresponds to a specific entity. Subsequently, we input the deterministic noise \( z \) and the prompt \( p \) into a pretrained teacher deterministic sampling diffusion model (e.g., BELM \cite{wang2025belm}), which produces a cross-attention map {\( A_i \)} at the final timestep $t$ to guide the calibration of our denoising network. 


\noindent \textbf{Calibration with LoRA.}
Our backbone network consists of an implicit text encoder and a denoising network UNet \(\epsilon_\phi\). The implicit text encoder is implemented by the CLIP image encoder \(\mathrm{CLIP_{image}}\) with an adapter. 
Formally, the extracted feature by our backbone can be written as:
\begin{equation}
  f_i = \epsilon_\phi({x_i},  \mathrm{MLP} \circ \mathrm{CLIP_{image}}({x_i})),  
  \label{eq:feat}
\end{equation}
where \(\mathrm{MLP}\) represents the learnable adapter applied to the CLIP image encoder and $x$ is the input image. 

Learning large-scale parameters for the UNet is challenging and could damage its preserved knowledge. Previous works \cite{smith2023continual,smith2023continual,gandikota2023concept} have shown that weight updates often have a low intrinsic rank when adapted for specific tasks. Therefore, we employ a lightweight approach, LoRA \cite{hu2021lora}, to fine-tune our UNet. The core of LoRA is to determine a set of weight modulations \(\Delta \mathbf{W}\) added to the original weight \(\mathbf{W}\).
For example, for the weights \(\mathbf{W}_k \in \mathbb{R}^{m_{in} \times m_{out}}\) in each cross-attention layer, we learn two smaller matrices \(\mathbf{B} \in \mathbb{R}^{m_{in} \times r}\) and \(\mathbf{D} \in \mathbb{R}^{r \times m_{out}}\), where \( r \) is the decomposition rank. The LoRA  updating is defined as:
\begin{equation}
   \mathbf{W}_k = \mathbf{W}_k + \Delta \mathbf{W}_k = \mathbf{W}_k + \mathbf{B} \times \mathbf{D}. 
\end{equation}
Similarly, we apply the same approach to learn the other projection layer \( \mathbf{W}_v \). 

\noindent \textbf{Calibration Loss.} At the finetuning, we use the training set of VG \cite{krishna2017visual} and PSG \cite{yang2022panoptic} to calibrate our denoising network respectively. Specifically, we feed the realistic images into our denoising network, resulting in the corresponding cross-attention map $A_i$. Then, we  align them with the  inversion cross-attention map from the pre-trained diffusion model as:
\begin{equation}
    \mathcal{L}_{cal} = \frac{1}{N} \sum_{i=1}^{N} \left(\lambda  \Vert A_i - A_i^\prime \Vert_1 \right),
\end{equation}
where we use the  L$_1$ norm loss on the cross-attention maps and $\lambda$ is a weighted factor. During the fine-tuning, we only update the $\operatorname{MLP}$ adapter of the $\operatorname{CLIP}$' image encoder and the lower rank matrices $\mathbf{B}$ and $\mathbf{D}$ of cross-attention layers.

\subsection{Spatial-aware Context Reasoning with RGT} \label{sec:4.2}

As discussed previously, current open-vocabulary PSG and SGG models primarily focus on designing visual prompts \cite{he2022towards,li2024pixels,zhou2024openpsg,chen2023expanding} to extract open-world knowledge from large multimodal models for novel relation prediction. However, these approaches often overlook the contextual information of relation pairs, particularly the spatial and semantic correlations among all entities within an image. This oversight limits their capability to handle complex scenes characterized by high spatial relation ambiguity.
To address this challenge, we propose a spatial-aware relation graph transformer for enhanced spatial and semantic context reasoning. Our method integrates graph convolutional networks with self-attention mechanisms to capture both long-range and local correlations among subjects and objects, whether they are directly related or not. This integration enables the generation of high-quality relation queries that better capture the underlying spatial and semantic structure of the scene.

\noindent \textbf{Spatial-semantic Graph Construction.}
The instance decoder \(\mathrm{\Phi}_{ins}\) takes the enhanced feature \( F \)  from the calibrated UNet \(\epsilon_\phi\) and instance queries \( Q_{o} \in \mathbb{R}^{N \times d} \) as input, {where \(N\) is the number of queries and \(d\) is the feature dimension}, producing a set of predicted masks \(\{m_i\}_{i=1}^N\) with corresponding labels.   The segmentation process is as follows: 
\begin{equation}
    \mathbf{H}_o = \Phi_{ins} (\mathbf{Q}_{o}, F),
\end{equation}
where $\mathbf{H}_o$ is the output object features.  Then, we construct a spatial-semantic graph  \( G \in \mathbb{R}^{N \times N} \) based on the spatial and semantic distances between instance pairs. The spatial distance is defined as \(1\) if two masks are neighboring and \(0\) otherwise. The semantic distance is set to \(1\) if the cosine similarity between the features of two objects (i.e., $\mathbf{H}_o$) exceeds a threshold and \(0\) otherwise.    The final graph \( G \) is formed by summing the two distances; if the resulting value is \(2\), it is capped at \(1\). 

\begin{figure}[t]
  \includegraphics[width=.42\textwidth]{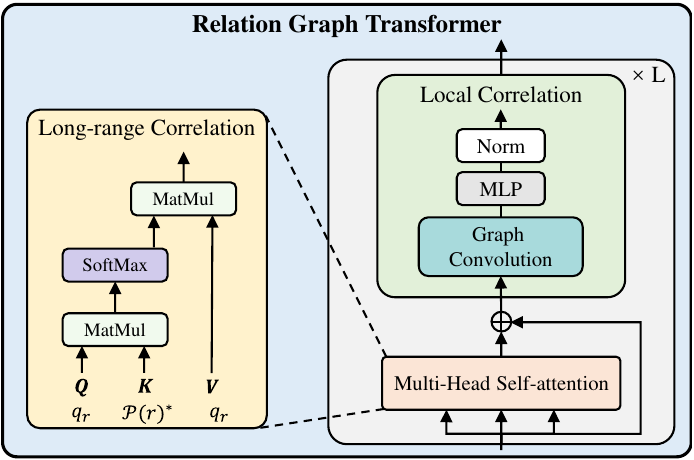}
  \centering
  \caption{Overview of the proposed Relation Graph Transformer,
which consists of long-range and local context reasoning components. It consists of $L$=$8$ identical blocks. 
  }
  \label{fig:rgt}
\end{figure}

\noindent \textbf{Relation Graph Transformer.} 
Intuitively, if two objects are connected in \( G \), there is a higher likelihood of a relationship between them thanks to the closer initial distance on the spatial and semantics. To further capture long-range correlations among objects, we propose a spatial-aware Relation Graph Transformer (RGT) that iteratively captures pairwise global long-range and local correlations by aggregating messages from both connected and non-connected objects. The RGT architecture is shown in Figure  \ref{fig:rgt}.

\noindent \textbf{Long-range Context Learning.}
In RGT, we design a long-range correlation reasoning block $\operatorname{RGT_g}(.)$ to learn global context.  
Formally, given an object \(\mathbf{q}_r \in \mathbf{H}_{o}\), we calculate its self-attention with its neighbors as:
\begin{equation}
   \underset{\mathcal{P}(r)^+}{\operatorname{RGT_g}(\mathbf{q}_r)} \!=\! \mathrm{softmax} \!\!\left[\frac{( \mathbf{W}_K \!\operatorname{\phi}[\mathcal{P}(r)^+])^\top \! (\mathbf{W}_Q \mathbf{q}_r)}{\sqrt{{\vert \mathcal{P}(r)^+ \vert}}}\right] \!\!\mathbf{W}_V \mathbf{q}_r, 
\end{equation}
where \(\mathcal{P}(r)^+\) denotes the neighbors of node \( r \);  \(\operatorname{\phi}[\mathcal{P}(r)^+]\) is a function that averages  all neighbors' features of node \( r \); and \(\mathbf{W}_K\), \(\mathbf{W}_Q\), \(\mathbf{W}_V\) are projection layers.
Meanwhile, we also calculate self-attention with non-neighbors as:
\begin{equation}
    \underset{\mathcal{P}(r)^-}{\operatorname{RGT_g}(\mathbf{q}_r)}\! =\! \mathrm{softmax} \!\!\left[\frac{( \mathbf{W}_K\! \operatorname{\phi}[\mathcal{P}(r)^-])^\top\!(\mathbf{W}_Q \mathbf{q}_r)}{\sqrt{{\vert \mathcal{P}(r)^- \vert}}}\right] \!\!\mathbf{W}_V \mathbf{q}_r,
\end{equation}
where \(\mathcal{P}(r)^-\) denotes non-neighbors of node \( r \).

The updated node feature \(\mathbf{q}_r\) is computed through an element-wise sum to capture information from both connected and non-connected relations:
\begin{equation}
  \mathbf{q}_r \leftarrow \mathbf{q}_r + \underset{\mathcal{P}(r)^+}{\operatorname{RGT}(\mathbf{q}_r)} + \underset{\mathcal{P}(r)^-}{\operatorname{RGT}(\mathbf{q}_r)}.  
\end{equation}
  Next, we use an MLP to fuse all features:
\begin{equation}
   \mathbf{q}_r^\prime \leftarrow \operatorname{MLP}[\mathbf{q}_r; \operatorname{\phi}[\mathcal{P}(r)^+]; \operatorname{\phi}[\mathcal{P}(r)^-]]. 
\end{equation}

\noindent \textbf{Local Context Learning.} In addition to global long-range relations, we propose a local relation learning block $\operatorname{RGT_l}(.)$ to capture  local correlations between object pairs. This process is defined as:
\begin{equation}
   \hat{\mathbf{q}}_r ={\operatorname{RGT_l}}(\mathbf{q}_r^\prime)=\operatorname{GCN}(G, \mathbf{q}_r^\prime; \mathbf{W}_l),
\end{equation}
where \(\operatorname{GCN}(.)\) is a graph convolution network and \(\mathbf{W}_l\) represents learnable parameters.

\noindent\textbf{Pairwise Relation Queries Construction.}
With the updated object features $\hat{\mathbf{q}}_r \in \hat{\mathbf{Q}}_r$ by the $\operatorname{RGT}$, we then construct the pairwise relation queries for predicate prediction by selecting the  close objects. Concretely, given the object feature $\hat{\mathbf{q}}_r$, the constructed relation queries $\Psi_r$ is written as:
\begin{equation}
    \Psi_r = \operatorname{Select}_{>} \{\operatorname{Dis}(\hat{\mathbf{q}}_r,\hat{\mathbf{Q}}_r), \eta\},
    \label{eq:rqc}
\end{equation}
where $\operatorname{Dis(.)}$ is a distance function, e.g. cosine distance, and $\operatorname{Select}_>\{.\}$ is a function to return the   
  objects with a similarity score greater than a threshhold $\eta$.   Then, we obtain all relation queries $\mathbf{Q}_{rel} \in \mathbb{R} ^ {M\times d}$, {where $M$ is the number of selected queries}. At training, we  add a auxiliary loss to optimize $\Psi_r$ by:
\begin{equation}
    \mathcal{L}_{rqc} = \left|\frac{\hat{\mathbf{Q}}_r \hat{\mathbf{Q}}_r^\top}{|\hat{\mathbf{Q}}_r||\hat{\mathbf{Q}}_r^\top|}   - \Psi^\prime \right|^2,
\end{equation}
where $ \Psi^\prime \!\! \in\! \!{ \{0,1\}}^{N\!\times \!N}$ is a ground truth   pair indicator matrix.  

\subsection{Open-vocabulary Relation Prediction}
\label{sec:4.3}


In this section, we introduce our proposed prompt-based open-world object and relation prediction.  We concatenate all selected object pairs by Eq.(\ref{eq:rqc}) to form relation queries $\mathbf{Q}_{rel} \in \mathbb{R}^{M \times d}$ for the relation decoder $\Phi_{rel}$. The relation prediction is formulated as:
\begin{equation}
    \mathbf{H}_r = \Phi_{rel} (\mathbf{Q}_{rel}, F).
\end{equation}

Following \cite{chen2023expanding}, we leverage a prompt-based classifier to calculate the object category distributions $\mathbf{P}_{o}$ and relation category distributions $\mathbf{P}_{r}$ formally,
\begin{equation} 
\mathbf{P}_{o} = \text{softmax}(\mathbf{H}_{o}^\top \cdot \tau/(\epsilon_t (\mathcal{T}_{o}))),
\label{eq:o}
\end{equation}
\begin{equation} 
\mathbf{P}_{r} = \text{softmax}(\mathbf{H}_{r}^\top \cdot \tau/(\epsilon_t (\mathcal{T}_{r}))),
\label{eq:r}
\end{equation}
where \(\mathbf{H}_{o}\) is the output feature by the instance decoder \(\Phi_{ins}\), \(\tau\) is a learnable temperature parameter, and \(\mathcal{T}_{o}\) and \(\mathcal{T}_{r}\) are the corresponding object and predicate category embeddings for the text encoder of CLIP, respectively. 

Except for the prediction score from the diffusion model, we further leverage the pre-trained pretrained text-image discriminative visual feature in the first calibration stage to enhance the open-vocabulary object and relation classification.  
Specifically, for instance prediction, we  extract the input image's feature by the implicit captioner  \( \mathrm{CLIP_{image}} (x)\) and then  compute a masked image feature that falls inside the predicted mask $m_i$ via a pool operation as
\begin{equation}
    \mathbf{H}_{o}^{'} =  \text{Pooling}(\mathrm{MLP}\circ \mathrm{CLIP_{image}}(x), m_i).
\end{equation}

As for the relation prediction, we first combine the masks of the subject and object into a holistic mask via an element addition operation $\odot$. Then, we calculate the pooled image feature of the pair by:
\begin{equation}
    \mathbf{H}_{r}^{'} =  \text{Pooling}(\mathrm{MLP}\circ \mathrm{CLIP_{image}}(x), m_s \odot m_o).
\end{equation}

 Then, we use the same way as \cref{eq:o} and \cref{eq:r} to calculate the new object score  $\mathbf{P}_o^\prime$ and relation score $\mathbf{P}_r^\prime$, respectively. Finally,  we leverage a geometric mean technique \cite{gu2021open,ding2022decoupling,xu2023open} to fuse the two scores. For example, the final object prediction score is computed  as: \(
     \mathbf{P}^o_\mathrm{final} = \mathbf{P}_o^\alpha \cdot \mathbf{P}_o^{\prime{(1-\alpha)}}\), where $\alpha$ is a fixed weighting factor. Similarly, the enhanced relation prediction score is:    \( \mathbf{P}^r_\mathrm{final} = \mathbf{P}_r^\alpha \cdot \mathbf{P}_r^{\prime{(1-\alpha)}}\).





\noindent \textbf{Training Loss.} In this stage, the overall training loss consists of three terms:  mask prediction loss $\mathcal{L}_{maks}$ as \cite{yang2022panoptic}, relation query construction loss $\mathcal{L}_{rqc}$ and relation Contrastive learning loss $\mathcal{L}_{rel}$ as \cite{yang2024open,xu2023open}.  The overall loss is written as:
\begin{equation} 
\mathcal{L} = \mathcal{L}_{\text{rel}} + \lambda_{\text{rqc}} \mathcal{L}_{rqc}+\lambda_{\text{mask}} L_{\text{mask}}.
\end{equation}
where  $\lambda_{\text{rqc}}$ and $\lambda_{\text{mask}}$ are     weighting hyperparameters.

\section{Experiments}
\subsection{Experimental Setup and Datasets}

\begin{table}[t]
\centering
\resizebox{\columnwidth}{!}{
\begin{tabular}{l|cc|cc}
\toprule

\multirow{2}{*}{Methods} & \multicolumn{2}{c|}{Close-Set} & \multicolumn{2}{c}{Open-Set (OvR)} \\ \cmidrule{2-5} 
 & R/mR50 & R/mR100 & R/mR50 & R/mR100 \\ \midrule
\multicolumn{5}{l}{Closed Methods} \\ \midrule
MOTIF\cite{zellers2018neural} & 21.7 / 9.6 & 22.0 / 9.7 & 1.6 / 0.6 & 2.0 / 1.1 \\
VCTree\cite{tang2019learning} & 22.1 / 10.2 & 22.5 / 10.2 & 1.7 / 0.9 & 2.3 / 1.3 \\
GPSNet\cite{lin2020gps} & 19.6 / 7.5 & 20.1 / 7.7 & 2.2 / 1.5 & 2.9 / 1.8 \\
PSGTR\cite{yang2022panoptic} & 34.4 / 20.8 & 36.3 / 22.1 & 3.1 / 1.9 & 3.3 / 2.0 \\
PSGFormer\cite{yang2022panoptic} & 19.6 / 17.0 & 20.1 / 17.6 & 2.2 / 1.9 & 3.1 / 2.7 \\
ADtrans\cite{li2024panoptic} & 29.6 / 29.7 & 30.0 / 30.0 & 2.4 / 2.0 & 3.6 / 3.4 \\
PairNet\cite{wang2024pair} & 35.6 / 28.5 & 39.6 / 30.6 & 3.2 / 2.5 & 6.5 / 5.1 \\
HiLo\cite{zhou2023hilo} & 40.7 / 30.3 & 43.0 / 33.1 & 3.7 / 2.8 & 4.8 / 3.6 \\
DSGG\cite{hayder2024dsgg} & 41.6 / 36.8 & 47.3 / 43.4 & 3.9 / 3.5 & 5.6 / 5.0 \\ \midrule
\multicolumn{5}{l}{Open PSG Methods} \\ \midrule
OvSGTR$^\dag$\cite{chen2023expanding} & 37.6 / 30.5 & 41.4 / 28.3 & 19.3 / 12.4 & 22.8 / 14.0 \\
PGSG\cite{li2024pixels} & 32.7 / 20.9 & 33.4 / 22.1 & 15.5 / 10.1 & 17.7 / 11.5 \\
OpenPSG$^\dag$\cite{zhou2024openpsg} & \underline{42.8} / \underline{38.9} & \underline{49.3} / \underline{47.5} & \underline{21.2} / \underline{19.8} & \underline{25.1} / \underline{21.4} \\ \midrule
\textbf{SPADE} & \textbf{45.1} / \textbf{41.2} & \textbf{54.3} / \textbf{51.7} & \textbf{26.7} / \textbf{23.3} & \textbf{31.8} / \textbf{25.8} \\

\bottomrule
\end{tabular}
}
\caption{Compared to the state-of-the-art PSG and SGG models on the  \textbf{PSG}  dataset in the close- and open-set scenarios.  $\dag$ represents the reproduced results based on their released code. }
\label{T:P}
\end{table}

\noindent\textbf{Datasets.}  We evaluate our model on two datasets: PSG \cite{yang2022panoptic} and Visual Genome (VG) \cite{krishna2017visual}. The PSG dataset, derived from COCO, comprises $48,749$ annotated images, with $46,563$ allocated for training and $2,186$ for testing. It includes $80$ object categories, $53$ ``stuff'' background categories, and $56$ relationship categories. In contrast, the VG dataset—widely used in SGG tasks—contains $108,077$ images with scene graph annotations. For our evaluation, we utilize the VG subset, which focuses on $150$ object categories and $50$ relation categories.

\noindent\textbf{Evaluation Tasks and Metrics.}
We adopt the scene graph detection (SGDET) protocol to evaluate all baseline models. We report recall (R) and mean recall (mR) on the test sets of VG  and PSG. Additionally,  we assess the model across different evaluation settings: closed-set and open-set. For the open-set, following prior work \cite{he2022towards,li2024pixels,zhou2024openpsg,chen2023expanding}, we split the relations and objects into base and novel categories in a $7$:$3$ ratio. Following \cite{chen2023expanding}, we report the open-set results on two scenarios: open-vocabulary relation (\textbf{OvR}) and open-vocabulary relation and object (\textbf{OvD+R}). Notably, for this open-set
setting, at the calibration, we only use images whose object and predicate categories arise in the seen group to
avoid data leakage problems. 


\noindent \textbf{Implementation details.}
We utilize a pre-trained diffusion model \cite{rombach2022high} as the diffusion backbone for calibration. Features are extracted from every three U-Net blocks, resized, and structured into a feature pyramid-like FPN.  By default, we set the time step for the diffusion process to $t=0$ for our denoising network. 
For the mask decoder, we follow the Mask2Former \cite{cheng2022masked} architecture. Notably, for a fair comparison with baseline models, we use the VG and PSG datasets for the calibration respectively, instead of combining them. 
Both object and interaction decoders are standard three-layer Transformer decoders.  The hyperparameter of $\alpha$,  $\eta$,  $\lambda_{rqc}$ and $\lambda_{mask}$ are set to  $0.34$, $0.65$, $0.6$ and $1$. During training, we apply the same data augmentation strategies as previous methods \cite{cheng2022masked,wang2024pair}. The parameters of the diffusion model and CLIP remain frozen throughout. The AdamW optimizer is used with an initial learning rate of $10^{-4}$ and a weight decay of $10^{-4}$. The model is trained for a total of $80$ epochs, with the learning rate reduced to $10^{-5}$ at the $60$-th epoch. All experiments are conducted on four A$100$ GPUs. 

\subsection{Main Results}

\begin{table}[t]
\centering
\resizebox{\columnwidth}{!}{
\begin{tabular}{l|cc|cc}
\toprule
\multirow{2}{*}{Methods} & \multicolumn{2}{c|}{Close-Set} & \multicolumn{2}{c}{Open-Set (OvR)} \\ \cmidrule{2-5} 
 & R/mR50 & R/mR100 & R/mR50 & R/mR100 \\ \midrule
\multicolumn{5}{l}{Closed Methods} \\ \midrule
IMP\cite{imp} & 25.5 / 4.1 & 30.7 / 7.9 & 1.1 / 0.5 & 1.5 / 0.9 \\
MOTIF\cite{zellers2018neural} & 32.5 / 6.6 & 36.8 / 7.9 & 2.2 / 1.3 & 2.9 / 1.7 \\
VCTree\cite{tang2019learning} & 31.9 / 6.4 & 36.0 / 7.3 & 1.3 / 0.4 & 2.4 / 1.8 \\
SGTR\cite{li2022sgtr} & 24.6 / 12.0 & 28.4 / 15.2 & 2.6 / 1.3 & 3.9 / 2.1 \\
SSR-CNN\cite{teng2022structured} & 23.3 / {17.9} & 26.5 / 21.4 & 2.8 / 2.1 & 5.2 / 4.2 \\
PE-Net\cite{Zheng_2023_CVPR} & 26.5 / 16.7 & 30.9 / 18.8 & 2.2 / 1.4 & 4.3 / 2.6 \\
EGTR\cite{im2024egtr} & 30.2 / 5.5 & 34.3 / 7.9 & 2.4 / 0.4 & 3.6 / 0.8 \\
DSGG\cite{hayder2024dsgg} & 32.9 / 13.0 & 38.5 / 17.3 & 3.5 / 1.4 & 5.2 / 2.3 \\ \midrule
\multicolumn{5}{l}{Open Methods} \\ \midrule
VS\(^3\)\cite{zhang2023learning} & {34.5} / ~—~ & 39.2 / ~—~ & 15.6 / 6.7 & {17.2} / 7.4 \\
OvSGTR$^\dag$\cite{chen2023expanding} & \underline{35.8} / 7.2 & \underline{41.3} / 8.8 & \underline{22.9} / 4.6 & \underline{26.7} / 5.7 \\ 
PGSG\cite{li2024pixels} & 20.3 / 10.5 & 23.6 / 12.7 & 15.8 / 8.2 & 19.1 / 10.3 \\
OpenPSG$^\dag$\cite{zhou2024openpsg} & 32.7 / \underline{13.5} & 39.0 / \underline{18.3} & 20.4 / \underline{9.4} & 25.7 / \underline{12.1} \\
\midrule
\textbf{SPADE} & \textbf{37.2} / \textbf{17.3} & \textbf{43.4} / \textbf{21.6} & \textbf{24.1} / \textbf{11.2} & \textbf{29.9} / \textbf{13.9} \\

\bottomrule
\end{tabular}
}
\caption{Compared to the state-of-the-art PSG and SGG models on the  VG  dataset in the close-set and open-set (OvR) scenarios. The second best results are underlined.  }
\label{T:S}
\end{table}

\begin{table}[t]
\centering
\resizebox{\columnwidth}{!}
{
\begin{tabular}{l|l|cccc}
\toprule
{Datasets} & {Methods} &  R@50 &mR@50 &R@100 & mR@100 \\ \midrule
\multirow{4}{*}{PSG} & PGSG\cite{li2024pixels} & 7.3 & 4.3 & 11.4 & 6.1 \\
 & OvSGTR$^\dag$\cite{chen2023expanding} & \underline{19.1} & \underline{13.4} & \underline{23.6}  & \underline{16.2} \\
 & OpenPSG$^\dag$\cite{zhou2024openpsg} & 11.4 &  6.7 & 16.3 &  7.7 \\ \cmidrule{2-6} 
 & \textbf{SPADE} & \textbf{22.7} & \textbf{17.8} &\textbf{ 26.2} & {20.1} \\ \midrule
\multirow{6}{*}{VG} & Cacao\cite{yu2023visually} & 1.2 &  0.5 & 3.3 &  1.4 \\
 & VS\(^3\)\cite{zhang2023learning} & 5.8&  2.5 & 7.2 &  3.1 \\
 & PGSG\cite{li2024pixels} & 5.3 &  2.8 & 8.1 &  4.4 \\
 & OpenPSG$^\dag$\cite{zhou2024openpsg} & 9.5 &  6.9 & 14.6 &  8.9 \\
 & OvSGTR\cite{chen2023expanding} & \underline{17.1} & \underline{10.4} & \underline{21.0} & \underline{14.8} \\ \cmidrule{2-6} 
 & \textbf{SPADE} & \textbf{19.4}& \textbf{13.1} & \textbf{24.7} &  \textbf{16.4} \\ \bottomrule
\end{tabular}}
\caption{Compared to the state-of-the-art PSG and SGG models on the  VG and PSG  dataset in the  open-set (OvD+R) \cite{chen2023expanding} scenario.   }
\label{T:ovdr}
\end{table}

\noindent \textbf{Results on PSG.}
We first present the results on the PSG dataset, as shown in Table \ref{T:P}, for both the closed- and open-set settings. For the open-set, we report results at two scenarios \cite{chen2023expanding}: OvR (training with full objects and partial predicates) and OvR+D (training with partial objects and predicates). The results for the OvR+D  on the PSG dataset are provided in Table \ref{T:ovdr}. To ensure a fair comparison, we employ the Mask2Former  \cite{cheng2022masked} with the Swin-B backbone \cite{liu2021swin} as the object segmenter across  PSG models.

The results indicate that, in the closed-set scenario, our proposed model, SPADE, achieves the highest performance on the PSG dataset. Specifically, SPADE also demonstrates superior performance in the open-set setting. Specifically, it outperforms OpenPSG \cite{zhou2024openpsg} by approximately $4$ points on average in the closed-set scenario and $3$ points in the open-set setting. Notably, under the OvR+D scenario, our model achieves a $5.5$\% improvement in R@$50$ compared to the competitive OpenPSG model \cite{zhou2024openpsg}. This significant performance gain highlights SPADE's strong capability in recognizing unseen objects and predicates.

\noindent \textbf{Results on VG.}
To evaluate our model on the VG dataset, we adapted the mask prediction heads of our instance decoder to box prediction heads, following the approach   \cite{zhou2024openpsg,li2024pixels}. The remaining settings are consistent with those used in OvSGTR \cite{chen2023expanding}. We report results for both close-set and open-set settings in Table \ref{T:S}. 
From the results, it shows that SPADE continues to achieve good performance on the VG dataset, surpassing many competitive open-set SGG models.  In the open-set, SPADE demonstrates significant superiority over other models, achieving   $\sim8$ points higher than OpenPSG in terms of OvD+R (R@$100$).

\begin{table}[t]
\centering
\resizebox{0.90\columnwidth}{!}{
\begin{tabular}{l|cc|cc}
\toprule
 \multicolumn{1}{c|}{\multirow{2}{*}{Methods}} & \multicolumn{2}{c}{NDR} & \multicolumn{2}{c}{DR} \\ 
\cmidrule(lr){2-3} \cmidrule(lr){4-5}  & R@50   & mR@50   & R@50   & mR@50   \\ \midrule
PGSG\cite{li2024pixels}              & 35.6 & 24.5 & 28.3 & 18.1 \\
OvSGTR\cite{chen2023expanding}       & 40.3 & 33.4 & 34.5 & 26.7 \\
OpenPSG\cite{zhou2024openpsg}        & 43.7 & 39.5 & 37.1 & 32.8 \\\midrule
\textbf{SPADE}  & \textbf{46.5} & \textbf{42.8} & \textbf{42.3} & \textbf{38.7} \\ \bottomrule
\end{tabular}
}
\caption{Distant and non-distant relations detection performance of VLM-based PSG models on the PSG dataset.}
\label{T:sp}
\vspace{-12pt}
\end{table}

\subsection{Ablation Study}
\label{sec:5.3}

In this section, we investigate four crucial parts of \textbf{SPADE}: {spatial relation prediction}, fine-tuning strategy,  the relation graph transformer and the open-vocabulary module. 


\noindent \textbf{Spatial Relation Prediction Analysis.} \label{spatial} To assess the spatial relation detection of VLM-based PSG models, we categorize relationships into spatial and non-spatial groups  and focus only on spatial relationships in this analysis.
We further define two evaluation scenarios: Distant Relation  (DR) and Non-Distant Relation Pairs (NDR). DR includes object pairs where the distance between the centers of their bounding boxes exceeds one-third of the image width, while NDRP includes object pairs where this distance is less than one-third. The results are shown in Table~\ref{T:sp}.  To ensure fairness, we use an equal number of DR and NDR samples during training to eliminate potential data bias.
From the results, we observe that many LMM-based models perform better in the NDR scenario but struggle in the DR scenario. For instance, OpenPSG shows an $7$\% performance gap in R@$50$ between DR and NDR predicates. In contrast, our model achieves more balanced performance across both groups, outperforming prior models in both distant and non-distant relationship prediction. This suggests that existing VLM-based models have difficulty capturing spatial relationships. 



\begin{table}[!t]
\centering
\resizebox{0.90\columnwidth}{!}{
\begin{tabular}{l|cc|cc}
\toprule
             \multirow{2}{*}{Strategies}                  & \multicolumn{2}{|c}{OvR}                            & \multicolumn{2}{c}{OvD+R}     \\  
   \cmidrule(lr){2-3} \cmidrule(lr){4-5}    &  {R@50}          &  {R@100}         &  {R@50}          &  {R@100}        \\ \midrule
{w/o~inversion}   & 21.0          & {25.3}          & 15.9          & 18.9          \\
{w/o~LoRA}        & 18.8          & {21.8}          & 12.7          & 15.8          \\
{w/o~calibration} & 15.3          & {18.9}          & 10.1          & 12.5          \\ \midrule
{\textbf{Ours}}   & \textbf{26.7} & \multicolumn{1}{c|}{\textbf{31.8}} & \textbf{22.7} & \textbf{26.2} \\ \bottomrule
\end{tabular}
}
\caption{Ablation results with different fine-tuning strategies for calibration on the open-set PSG dataset.}
\label{T:ti}
\end{table}

\noindent\textbf{Calibration Strategies.} We examine different fine-tuning techniques: without inversion sampling (w/o~inversion), i.e, random sampling a noise from a  Gaussian distribution; fine-tuning all UNet parameters (w/o~LoRA); and using the pre-trained UNet directly (w/o~calibration). The results are presented in Table \ref{T:ti}.
From the results, we observe that w/o calibration yields the worst performance, confirming that the knowledge embedded in the pre-trained diffusion model differs significantly from that required for scene graph understanding. This highlights the necessity of our calibration operation.  
Similarly, w/o LoRA also performs poorly, likely because modifying all parameters of the denoising network disrupts the world knowledge preserved in the pre-trained diffusion model. Additionally, using a random sampling diffusion model does not yield strong performance, probably because the generated image features and spatial layout are inconsistent with the input real image.
In contrast, our approach—LoRA with a deterministic sampling model—preserves the pre-trained parameters while learning two low-rank matrices specifically tailored for the PSG task. This provides a more effective fine-tuning strategy, balancing knowledge retention and task-specific adaptation.


\begin{table}[t]
\centering
\resizebox{0.97\columnwidth}{!}{
\begin{tabular}{cccc|cc}
\toprule



LCNL       & LCNNL      & LCL        & \(\mathcal{L}_{rqc}\) & R/mR@50     & R/mR@100    \\ \midrule
           &            &            &                       & 30.5 / 26.3 & 40.2 / 35.7 \\
\checkmark &            &            &                       & 35.6 / 32.2 & 46.5 / 43.7 \\
\checkmark & \checkmark &            &                       & 37.1 / 34.4 & 47.9 / 44.8 \\
\checkmark & \checkmark & \checkmark &                       & 40.3 / 35.6 & 50.8 / 46.5 \\ \midrule
\checkmark & \checkmark & \checkmark & \checkmark & \textbf{45.1} / \textbf{41.2} & \textbf{54.3} / \textbf{51.7} \\

\bottomrule
\end{tabular}
}
\caption{Ablation study results of the Relation Graph Transformer on the close-set PSG dataset. 
}
\label{tab:LCL}
\end{table}

\begin{table}[t]
\centering
\resizebox{0.97\columnwidth}{!}{
\begin{tabular}{cc|cc|cc}
\toprule
 \multirow{2}{*}{Diffusion} & \multirow{2}{*}{Pooling}  & \multicolumn{2}{|c}{OvR} & \multicolumn{2}{c}{OvR+D} \\ \cmidrule(lr){3-4} \cmidrule(lr){5-6} 
 &  & R@50 & R@100 & R@50& R@100 \\ \midrule
 & \checkmark & 12.5 & 14.6 & 8.5 & 9.2 \\
\checkmark &  & 21.4 & 24.8 & 12.8 & 14.7 \\ \midrule
\checkmark & {\checkmark} & \textbf{26.7} & \multicolumn{1}{c|}{\textbf{31.8}} & \textbf{22.7} & \textbf{26.2} \\ \bottomrule
\end{tabular}
}
\caption{Ablation results with  open-vocabulary  classification strategies on the   PSG dataset in the open-set setting.}
\label{T:ov}
\end{table}

\noindent \textbf{Relation Graph Transformer.}  
In this section, we analyze three key components of the relation graph transformer: {long-range correlation neighbors learning (LCNL), long-range correlation non-neighbors learning (LCNNL), and local correlation learning (LCL).} Additionally, we examine the impact of the auxiliary loss 
  \(\mathcal{L}_{rqc}\).
Table~\ref{tab:LCL} presents the ablation results, highlighting the contributions of each component. The baseline model, which excludes all components, achieves the lowest performance. Adding LCNL or LCNNL individually leads to noticeable improvements, while incorporating LCL further enhances the results. Notably, the full configuration—integrating LCNL, LCNNL, LCL, and 
 \(\mathcal{L}_{rqc}\)
 —achieves the highest scores.
These findings confirm that both long-range (via neighbor and non-neighbor correlations) and local relation modeling, along with auxiliary relation query construction loss, are essential for robust relation prediction.

\noindent \textbf{Open-Vocabulary Module.}  
Table~\ref{T:ov} presents the ablation results to assess the impact of incorporating diffusion features and pooled features on open-vocabulary relation prediction. The results indicate that integrating pooled features leads to notable improvements over the baseline, which relies solely on the predictions of the diffusion model. This suggests that diffusion-based representations alone may not be sufficient for predicting open-vocabulary relations and that pooled features provide complementary information that enhances the model’s predictive capabilities. These findings highlight the importance of both diffusion-based features and pooled features for effective open-vocabulary relation prediction.

\begin{figure}[t]

  \includegraphics[width=.5\textwidth]{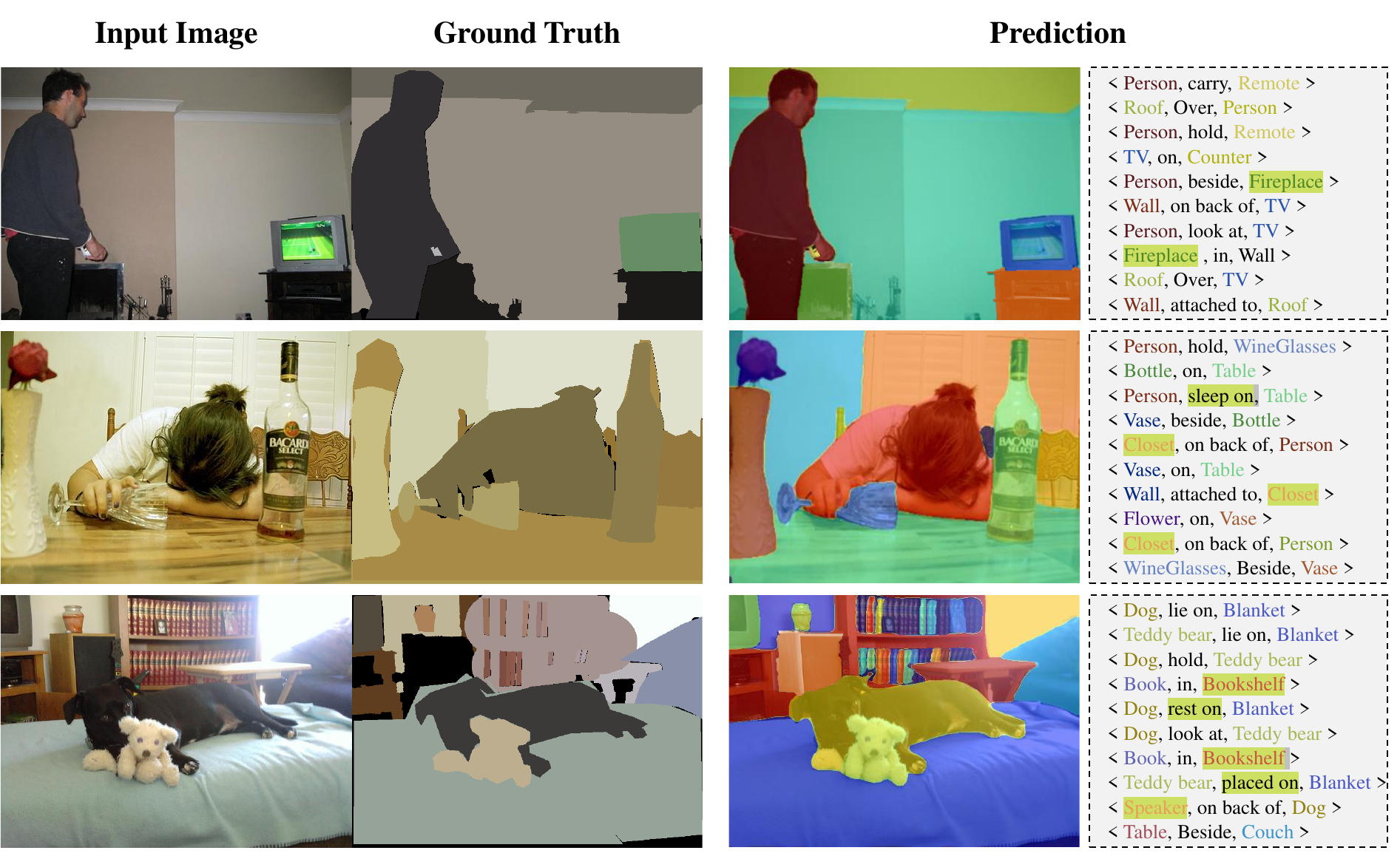}
  \centering
  \caption{
  The visualization results of  PSG by \textbf{SPADE}. The first column shows the input images, while the second column presents the ground truth. The other columns display the predictions by our model, where unseen predicted objects and relationships are highlighted on a yellow green background.
  }
  \label{fig:vis2}
\end{figure}

\subsection{Visualization Results}


 In Figure~\ref{fig:vis2}, we present visualization results that illustrate our model’s open-vocabulary capabilities.  These visualizations reveal that our model not only successfully identifies and segments objects that are entirely unseen during training (highlighted in yellow-green) but also adeptly infers novel relationships among them.  This indicates that SPADE is capable of generalizing beyond the set of categories encountered during training, effectively capturing both known and unknown object and predicate classes. 



\subsection{Conclusion}
In this paper, we introduced SPADE, a novel spatial-aware diffusion-based framework for open-vocabulary PSG. SPADE addresses the limitations of VLM-based PSG models, particularly their weaknesses in spatial relation reasoning.  
Our approach consists of two key steps: inversion-guided calibration and spatial-aware context reasoning. First, we fine-tune a pre-trained teacher diffusion model into a PSG-specific denoising network using cross-attention maps from inversion, optimized with a lightweight LoRA-based calibration strategy. Second, we introduce a spatial-aware relation graph transformer that captures both local and long-range contextual dependencies, improving relation query generation.  
Extensive experiments on benchmark PSG and Visual Genome datasets demonstrate that SPADE achieves state-of-the-art performance in both closed-set and open-set scenarios.
For future work, we will extend its applicability to broader multimodal tasks, e.g., human-object interaction detection.

\section*{Acknowledgments}

This research was partially supported by the National
Natural Science Foundation of China (NSFC) (granted No. 62306064), the  Central-Guided Local Science and Technology Development  (granted No. 2023ZYD0165), and Sichuan Science and Technology Program (granted No. 2024ZDZX0011). We appreciate all the authors for their
fruitful discussions. In addition, thanks are extended to
anonymous reviewers for their insightful comments and
suggestions.
{
    \small
    \bibliographystyle{ieeenat_fullname}
    \bibliography{main}
}

\end{document}